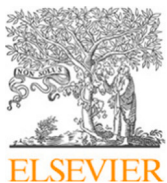
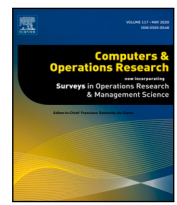

# Variable selection for Naïve Bayes classification

Rafael Blanquero [a,c], Emilio Carrizosa [a,c], Pepa Ramírez-Cobo [b,c], M. Remedios Sillero-Denamiel [a,c,*]

[a] *Departamento de Estadística e Investigación Operativa, Universidad de Sevilla, Seville, Spain*
[b] *Departamento de Estadística e Investigación Operativa, Universidad de Cádiz, Cádiz, Spain*
[c] *IMUS, Instituto de Matemáticas de la Universidad de Sevilla, Seville, Spain*

A B S T R A C T

The Naïve Bayes has proven to be a tractable and efficient method for classification in multivariate analysis. However, features are usually correlated, a fact that violates the Naïve Bayes' assumption of conditional independence, and may deteriorate the method's performance. Moreover, datasets are often characterized by a large number of features, which may complicate the interpretation of the results as well as slow down the method's execution.

In this paper we propose a sparse version of the Naïve Bayes classifier that is characterized by three properties. First, the sparsity is achieved taking into account the correlation structure of the covariates. Second, different performance measures can be used to guide the selection of features. Third, performance constraints on groups of higher interest can be included. Our proposal leads to a smart search, which yields competitive running times, whereas the flexibility in terms of performance measure for classification is integrated. Our findings show that, when compared against well-referenced feature selection approaches, the proposed sparse Naïve Bayes obtains competitive results regarding accuracy, sparsity and running times for balanced datasets. In the case of datasets with unbalanced (or with different importance) classes, a better compromise between classification rates for the different classes is achieved.

## 1. Introduction

Among the assortment of current classification techniques, the Naïve Bayes (NB) classifier has played a prominent role because of its simplicity, tractability and efficiency, see Hand and Yu (2001). The implicit assumption of independent features conditioned to the class eases the NB implementation significantly, since it allows the decomposition of a sample likelihood into a product of univariate marginals. In addition, the NB usually estimates fewer parameters than other renowned classifiers, so it is less prone to overfitting (Domingos and Pazzani, 1997; Hand and Yu, 2001). As a consequence, a number of applications of the NB in real contexts can be found, for example, in medicine (Wolfson et al., 2015), genetics (Minnier et al., 2015), reliability (Turhan and Bener, 2009), risk (Minnier et al., 2015) or document analysis (Guan et al., 2014), among others. Nowadays, datasets are usually characterized by a large number of features and, although such high dimensionality does not represent a major computational drawback when running the NB, it may have negative consequences in terms of the comprehensibility of its solutions (Carrizosa and Romero Morales, 2013). The search for more interpretable solutions, also common in other multivariate contexts such as regression (Cai et al., 2009; Carrizosa et al., 2021; Lin et al., 2011), clustering (Benati and García, 2014; Maldonado et al., 2015), time series analysis (Blanquero et al., 2020; Carrizosa et al., 2017) or visualization (Carrizosa and Guerrero, 2014), has led to the development of sparse multivariate techniques, see Hastie et al. (2015). Sparsity in classification is closely linked to the concepts of *Variable Selection* and *Feature Selection* (Carrizosa et al., 2016; George and McCulloch, 1993; Lin et al., 2011; Zou and Hastie, 2005), whose aim is to identify the relevant variables within a set of many predictors so that classification accuracy is not reduced.

In this paper, we propose an alternative sparse method for databases with dependent features. In particular, we embed a variable reduction algorithm within the NB's scheme to produce a sparse version of the classifier. Our aim is two-fold: on one hand, sparsity is pursued in the sense that only a subset of predictive features is used by the classifier's construction, making the so-obtained classifier more interpretable, and, on the other hand, we have a flexible framework to chose the accuracy measure to be optimized so that the classifier's performance does not worsen with respect to the classic NB.

\* Corresponding author.
 *E-mail address:* rsillero@us.es (M.R. Sillero-Denamiel).

Some works have addressed different strategies for variable reduction for the NB. For example, Feng et al. (2015) and McCallum and Nigam (1998) base their feature selection approaches on the univariate correlations between features and the class. In this sense, Chen et al. (2020), Tang et al. (2016a) and Tang et al. (2016b) aim to rank the features according to their capacity for classification or a specific feature selection criterion. In Zhang et al. (2009), the use of the principal components technique and genetic algorithms to remove irrelevant and redundant features are examined. The Evolutional Naïve Bayes (Jiang et al., 2005) is a wrapper which also performs a genetic search to select a subset from the whole set, although it is sensitive to many parameters, which is disadvantageous in practice. Other studies which are also focused on *hard variable selection* approaches to reduce the number of redundant predictors are Bermejo et al. (2014) and (Mukherjee and Sharma, 2012). In this sense, Langley and Sage (1994) define the *selective Naïve Bayes* (SNB) classifier, which is based on a wrapper approach (Kohavi and John, 1997). However, due to the complexity of the involved search algorithm and its tendency to make overfitting, the SNB does not perform well on large datasets (Boullé, 2007). Therefore, a Bayesian approach – defined as SNB(MAP) – is considered in Boullé (2007) to improve the performance of the SNB so that a compromise between the performance of the classifier and the sparsity is found. Another example can be found in "ann" Ratanamahatana and Gunopulos (2003), which proposes a method that combines NB and decision trees.

However, as pointed out by Boullé (2007), it is important to "*exploit multivariate preprocessing methods in order to circumvent the Naïve Bayes assumption*". In this paper, we adopt this scheme and propose a *hard variable selection* process which is motivated by the conditional independence assumption of the NB. It is known that the NB is Bayes-optimal (that is, it guarantees the minimum classification error), when the predictors are independent conditioned to the class (Kuncheva, 2006). On the other hand, it is also well documented in the literature that conditional independence is a sufficient condition but not necessary to get the optimal NB (Domingos and Pazzani, 1997; Hand and Yu, 2001; Hastie et al., 2001). Even if the fact that features are conditionally independent might not make a significant difference with respect to the situation where features are correlated, such slight difference in the NB performance may be crucial for some real contexts (cancer diagnosis, for example). The sparse version of the NB proposed in this paper, which is suitable for dealing with correlated patterns in datasets, is obtained by integrating a variable reduction method in such a way that only certain combinations of features, chosen according to their degree of dependence, are considered. Other papers have considered before correlations among the features as is the case of Hall (2000), Jiang et al. (2019) and Rezaei et al. (2018). The filter *Correlation based Feature Selection* (CFS) (Hall, 2000) is based on the assumption that a good subset of attributes should be highly correlated with the response variable but, on the other hand, there should exist few dependencies among them. This hypothesis is also used in Jiang et al. (2019), where a correlation-based feature weighting filter for NB is developed. In Rezaei et al. (2018), clustering is used to detect groups of correlated features and select only a small number of attributes. In particular, the optimal number of clusters stems from the mean silhouette score, which measures how similar a variable is to its own cluster compared to other clusters.

Additionally, the novel strategy can be implemented using the most adequate performance measure given the properties of the datasets. Minimizing the overall misclassification rate is always an option, but, for example, if datasets are unbalanced, the AUC (area under the ROC curve) may be preferred, since it is sensitive to class imbalance and, therefore, achieves a better compromise among the correct classification rates for the different classes. Recent works have considered different alternative performance measures, Jiang et al. (2012, 2019) and Zhang et al. (2020). For instance, the Randomly Selected Naïve Bayes (Jiang et al., 2012) considers the classification accuracy (ACC), AUC or conditional log likelihood; whereas in Jiang et al. (2019) and Zhang et al. (2020), two class-specific attribute weighted Naïve Bayes versions are defined.

Not only our method establishes the sparsity in terms of the correlation among the covariates and is flexible so that the most convenient classification measure can be used, but also it is a cost-sensitive classifier. When dealing with real-world applications where there exist groups at risk (as it happens in medical contexts, risk management, credit card fraud detection or when fair classification is a requirement as a social criterion), cost-sensitive learning approaches that assign different importance to the different groups should be considered (Leevy et al., 2018). Moreover, these methods turn out to be very convenient for unbalanced datasets, where the minority class may be the worst classified (and the most critical one). In particular, the inclusion of constraints on the proportions of correctly classified instances of such groups may be convenient for having direct control over their misclassification rates and obtaining adequate results for them (Benítez-Peña et al., 2019; Blanquero et al., 2021a,b). That is, whereas the global performance criterion is optimized, further control can be added via performance constraints on the groups of interest in each case. As it will be detailed, the sparse NB defined in this work is able to integrate such performance constraints.

This paper is organized as follows. In Section 2, a brief review of the NB is done, the notation is introduced, and some performance measures typically used in classification are reviewed. A numerical example motivating our approach for a sparse NB is presented next. In Section 3, the proposed version of sparse NB is described. Section 4 illustrates the new sparse classifier. Synthetic datasets as well as ten well documented real databases with different properties will be thoroughly analyzed, considering different performance measures and/or adding performance constraints in groups of interest. A complete discussion concerning the performance results, sparsity and running times of the proposed methodology in comparison with benchmark approaches will be given. Finally, some conclusions to this work and further related research are described in Section 5. Further information concerning the properties of the considered datasets and the choice of the tuning parameters will be described at the Supplementary Material.

## 2. Preliminaries

### 2.1. The Naïve Bayes classifier and performance measures

Consider a classification problem with a set of $p$ features $(X_1, \ldots, X_p)$ and $K$ possible classes. Given a new observation $\mathbf{x} = (x_1, \ldots, x_p)$, the aim is to assign $\mathbf{x}$ to one of the $K$ classes. The NB computes the conditional probabilities $p(C_k \mid \mathbf{x})$ for $k = 1, \ldots, K$ and $\mathbf{x}$ is assigned to the class $\hat{y} \in \{1, \ldots, K\}$ satisfying

$$\hat{y} = \operatorname*{argmax}_{k \in \{1,\ldots,K\}} p(C_k \mid \mathbf{x}).$$

The computation of $p(C_k \mid \mathbf{x})$ may be cumbersome if the number of features $p$ is large. However, the use of the Bayes theorem eases the previous computation since

$$p(C_k \mid \mathbf{x}) = \frac{\pi(C_k) p(\mathbf{x} \mid C_k)}{p(\mathbf{x})},$$

where $\pi(C_k)$ is the prior distribution for the class, $p(\mathbf{x} \mid C_k)$ is the likelihood function of the data and $p(\mathbf{x})$ is the so-called evidence. Since the evidence is the same for all the classes, in practice, the interest is in computing the numerator.

The key assumption of the NB is the independence of the features conditioned to the class, which implies that

$$p(\mathbf{x} \mid C_k) = p(x_1, \ldots, x_p \mid C_k) = p(x_1 \mid C_k) \ldots p(x_p \mid C_k) \tag{1}$$

and therefore, the probabilities of interest $p(C_k \mid \mathbf{x})$ are computed in a straightforward manner as proportional to (1). Note that, in (1), a probability distribution for the features conditioned to the class $X_i \mid C_k$ needs to be chosen by the user and estimated by some statistical method as, for example, a maximum likelihood criterion.

Several measures can be used to study a classifier's performance, see for example (Sokolova and Lapalme, 2009). In real contexts, besides good overall classification rates, high classification rates for specific classes may be sought. For this reason, throughout this work, we shall consider the classic *Recall of each class k* ($Recall_k$) for $k = 1, \ldots, K$, and also, the *accuracy (ACC)* and the *precision*, which are defined as follows,

$$Recall_k = \frac{(True\ Class\ k) \times 100}{Number\ of\ individuals\ in\ class\ k}, \tag{2}$$

$$ACC = \frac{\left(\sum_k True\ Class\ k\right) \times 100}{Total\ number\ of\ individuals}, \tag{3}$$

$$precision_k = \frac{True\ Class\ k}{(True\ Class\ k) + (False\ Class\ k)}, \tag{4}$$

as well as the AUC.

### 2.2. The independence assumption: a numerical example

The effect of the independence assumption over the performance of the NB when correlated features are analyzed, has been studied in the literature, see Domingos and Pazzani (1996, 1997), Hand and Yu (2001), Hastie et al. (2001) and Zhang (2004). As commented in Section 1, the conclusion is that, even though the independence assumption is not satisfied, the classifier's performance may not be considerably altered. However, using just a properly chosen subset of the variables may make the independence assumption less violated, and the accuracy improved (on top of the fact that a model with less variables is more explainable).

In order to illustrate how the violation of the independence assumption may affect the performance of the NB, consider the next numerical example. A sample of size 2000 of a random vector $(X_1, X_2, X_3, X_4)$ is simulated for two classes from a multivariate Normal distribution in such a way that the random variables are independent conditioned to the classes except for $X_1$ and $X_2$ which are correlated according to a Pearson coefficient of 0.95. A Gaussian NB classifier (that is, $X_i \mid C_k \sim N\left(\mu_{i,k}, \sigma_{i,k}^2\right)$ for $i = 1, 2, 3, 4$, $k = 1, 2$, where $\mu_{i,k}$ were randomly selected in the interval [1,7]) was run using all possible subsets of features and the results are shown in Table 1. The accuracy when all the variables are used is equal to 78.28, a value that is improved if the set $\{X_1, X_3, X_4\}$ is considered (accuracy equal to 79.94).

Having illustrated that using just a subset of the features may improve accuracy, we face the combinatorial problem of finding the adequate set of features to be used. The previous *brute force* procedure, where all possible combinations of features are examined, turns out infeasible in practice, especially for large databases. Instead, in this paper we propose a variable reduction method in which only certain combinations of features are sampled and evaluated. Such combinations, as will be seen in Section 3, shall be chosen by considering the dependencies among the features.

## 3. A sparse Naïve Bayes

As commented in Section 2.2, considering all possible combinations of features to determine the best one is hard from a computational point of view, especially for large datasets since a total of $2^p - 1$ sets should be evaluated. The aim of this section is to describe an efficient methodology to guide the search of the subset of features, by inspecting only some subsets selected in terms of the dependence among features. As a result, a sparse, computationally tractable NB is obtained.

### 3.1. Description of the method

The variable reduction strategy proposed in this section is based on a clustering of features made in terms of their dependencies. As commented in Section 2.1, the key assumption of the Naïve Bayes is the independence of the features conditioned to the class. The novel method presented in this work aims to preserve the independence assumption without damaging the predictive power of the classic NB. In other words, our methodology helps to select variables that are as independent as possible while provides good classification accuracy. To do that, we consider a dependence measure between random variables $X$ and $Y$, which increases with the degree of dependence between the variables. First, consider for $i, j \in \{1, \ldots, p\}$ and $k = 1, \ldots, K$, the dependence between feature $X_i$ and feature $X_j$ conditioned on class $C_k$. In order to have a unique, summarized measure of dependence between $X_i$ and $X_j$, let $M$ be the matrix whose elements $(M(i,j))$ represent the maximum dependence among all classes, between $X_i$ and $X_j$. Note that such a choice represents the worst case scenario. A number of dependence measures proposed in the literature can be selected: Pearson correlation coefficient, Spearman's rank-order correlation coefficient, Hoeffding $D$ statistic (see Hoeffding, 1948), the mutual information coefficient (MI) (Linfoot, 1957), the Maximal Information coefficient (MIC) (Reshef et al., 2011) or the distance correlation coefficient (Székely et al., 2007), among others. We tried using these different measures and similar results were obtained (see Section 4 and the Supplementary Material). Therefore, since the mutual information measure enables us to work with both continuous and categorical variables and has been widely used in the literature (Kinney et al., 2010; Sharpee et al., 2004), we will select this measure. This coefficient quantifies the information about one variable $X$ provided by a different variable $Y$, and it is defined as

$$I(X, Y) = \int_Y \int_X p(x, y) \log\left(\frac{p(x, y)}{p(x)p(y)}\right) dx dy,$$

in the case of continuous variables. In the categorical case, the previous formula can be rewritten in terms of sums. The previous dependence measure can be computed by the function `mutinformation` from the `infotheo` package of the Statistical software environment R (R Core Team, 2017). An illustration of the matrix $M$ for the real dataset *Statlog (Australian Credit Approval)* from the UCI Machine Learning Repository (Lichman, 2013) is represented by Fig. 1. The dataset concerns credit card applications, and is formed by 14 variables and two classes (+/-). Moreover, to visualize the different correlation patterns of the real-life datasets used throughout this work, in Section 3 of the Supplementary Material, the associated matrices $M$ using the MI measure are represented via heatmaps.

Next, with the aim of performing a cluster analysis in terms of the degree of association among the features, a dissimilarity matrix $H$ of dimension $p \times p$ is defined in terms of matrix $M$ by the elements

$$H(i, j) = \frac{M^* - M(i, j)}{M^*}, \tag{5}$$

where

$$M^* = \max_{i,j \in \{1, \ldots, p\}} M(i, j).$$

Note that, under the previous definition, the elements of $H$ are bounded below by zero, where this value represents the maximum degree of dependence. Moreover, the upper-bound of the elements of $H$ is one, which represents the minimum degree of dependence. Therefore, the higher the values of $H(i, j)$ are, the less dependence exists between $X_i$ and $X_j$, according to the selected dependence measure.

Note that, as described in Section 1 of the Supplementary Material, the results obtained are rather robust regarding the dependence measure. Once the dependence measure is set, the classifier's performance measure to be maximized in the embedded Variable Selection strategy can be chosen, among the previously described measures in Section 2.1, according to the user's convenience and the properties of the dataset. Generally, the ACC is selected, but in some cases, e.g. for unbalanced datasets or when there exist critical classes, our proposal replace ACC with AUC, *precision* or a certain *Recall*. Thus, the novel method turns out specially advisable for datasets where the classes are unbalanced and/or of different importance. The selection of the dependence and the classifier's performance measures is the first step of our algorithm (see Algorithm 1).

**Table 1**
Performance rate for all possible combinations of features in a multivariate normal simulated example.

| Combination of variables | $X_1$ | $X_2$ | $X_3$ | $X_4$ | $X_1,X_2$ | $X_1,X_3$ | $X_1,X_4$ | $X_2,X_3$ | $X_2,X_4$ | $X_3,X_4$ | $X_1,X_2,X_3$ | $X_1,X_2,X_4$ | $X_1,X_3,X_4$ | $X_2,X_3,X_4$ | $X_1,X_2,X_3,X_4$ |
|---|---|---|---|---|---|---|---|---|---|---|---|---|---|---|---|
| ACC | 68.08 | 68.51 | 68.55 | 69.01 | 68.38 | 74.99 | 75.08 | 74.86 | 75.25 | 75.68 | 73.58 | 73.85 | 79.94 | 79.84 | 78.28 |

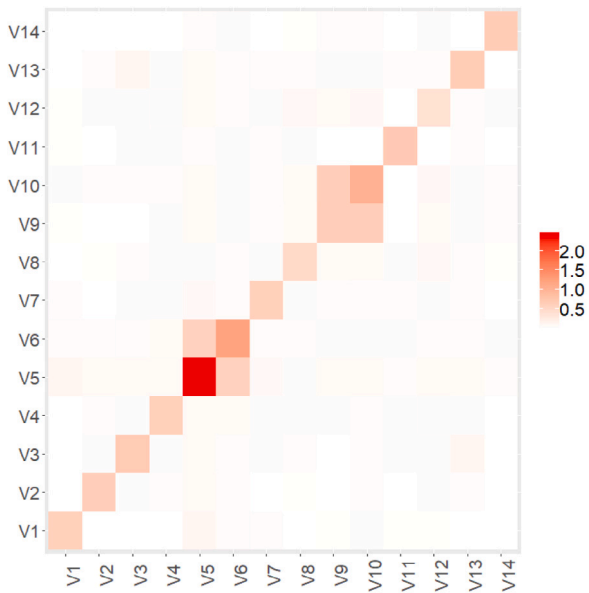

**Fig. 1.** Heatmap associated to matrix $M$ (based on MI correlation) corresponding to the *Australian* dataset.

Once we have chosen a dependence measure, and the elements of the matrix $H$ are computed, we perform a hierarchical cluster analysis of features according to the dissimilarity matrix $H$ (step 2 of the algorithm).

In the obtained dendrogram, the vertical axis represents the degree of dissimilarity. The higher the value of the height is, the less dependent the variables are, according to the dependence measure. For example, the dendrogram corresponding to the *Australian* dataset is given by Fig. 2. Such a dendrogram has been obtained using the routine `hclust` of the Statistical software environment R. In this case, $V_5$ and $V_6$ are highly dependent, as well as $V_9$ and $V_{10}$. However, the rest of variables are almost independent, since they cluster at heights between 0.9 and 1.

Once the dendrogram is built, a (not necessarily regular) grid of a specified number $C$ of cuts along the height is fixed. The basic idea underlying the variable reduction strategy is to examine at each cut (or threshold) of the grid several combinations of features, in such a way that only one feature is selected per cluster since all elements in a cluster are assumed to be strongly dependent. As an example, consider Fig. 2 and assume that one of the $C$ cuts is $c = 0.74$ (horizontal line). Then, we consider that there are 12 clusters: 10 clusters formed by only one feature and the clusters $\{V_5,V_6\}$ and $\{V_9,V_{10}\}$. And, therefore, four independent combinations would be selected at this threshold: $(V_1,V_7,V_{11},V_3,V_{13},V_2,\boldsymbol{V_5},V_4,V_{14},V_8,V_{12},\boldsymbol{V_9})$, $(V_1,V_7,V_{11},V_3,V_{13},V_2,\boldsymbol{V_5},V_4,V_{14},V_8,V_{12},\boldsymbol{V_{10}})$, $(V_1,V_7,V_{11},V_3,V_{13},V_2,\boldsymbol{V_6},V_4,V_{14},V_8,V_{12},\boldsymbol{V_9})$ and $(V_1,V_7,V_{11},V_3,V_{13},V_2,\boldsymbol{V_6},V_4,V_{14},V_8,V_{12},$

$\boldsymbol{V_{10}})$. Note that the higher (lower) the value of the cut, the more likely we are to choose independent (dependent) variables.

Although the previous strategy reduces the computational cost of the *brute force* approach, it still may be costly for large datasets that originate a complex dendrogram with many combinations per threshold. In addition, removing some of the features from the combinations may lead up to sparser and more accurate models, since it might happen that the (independent) variables selected in the combinations have a very low predictive power. In order to strive to avoid such inconveniences, we propose a refinement of the strategy as follows. First, a maximum number $S$ of combinations per threshold is set (if the total number of possibilities for a given threshold $c$, $nc(c)$, does not exceed $S$, then all of them will be considered). In the previous example, $nc(0.74) = 4$. We should point out that parameters $C$ and $S$ are used to alleviate the computational burden, since, as commented before, $C$ fixes the number of cuts along the height in the dendrogram, and $S$ the maximum number of combinations per threshold to be evaluated. Therefore, the higher $C$ and $S$ are, the higher the computer time is. For this reason, we will fix reasonable values for these parameters in Section 4.3. Second, for each cluster of variables to be examined, a value $q$ representing the probability of selecting this cluster for extracting randomly a variable to be included in the combination is also set. If we fix $q = 0.4$, the previous four combinations become $(V_1,V_7,V_8,V_{13},V_{14})$, $(V_1,V_4,V_6,V_8,V_9,V_{12},V_{14})$, $(V_2,V_8)$ and $(V_4,V_6,V_{10},V_{11})$, respectively. The parameter $q$ is directly related to the sparsity degree: the lower the value of $q$ is, the less variables are inspected (the expected number variables to be considered is equal to $q \times p$). The choice of the values $\{C,S,q\}$ will be discussed in Section 4.

Once the set of combinations of features to be evaluated is reduced, the NB would be implemented and, its performance and feasibility on the constrains considered, evaluated for each combination. This is summarized in step 3 of the Algorithm 1.

Finally, the feasible combination yielding the highest performance measure (accuracy, AUC or whatever chosen measure) would be considered the best, taking also into account the whole set of variables in this comparison (step 4). For the *Australian* database example, if no constraint is imposed, the features selected by our model are $(V_1,V_4,V_6,V_8,V_9,V_{12},V_{14})$, which achieve an ACC of 86.76, whereas the whole set of variables returns 85.29. According to the results, it can be deduced that our model has kept the important features, using only a half of the total set. However, in this dataset, the positive class (the load is granted) is the most risky. Then, if we impose e.g. that Recall + > 92, the combination of features $(V_2,V_8)$ would be the selected one.

A summary concerning the strategy for the sparse NB is given by Algorithm 1.

## 4. Numerical illustrations

In this section, the behavior and performance of our approach is illustrated throughout an extensive empirical study, using both simulated and real datasets. In the first case, a synthetic data set is simulated in order to test how the performance and level of sparsity of the proposed sparse NB changes with the level of dependence among the features. Second, ten real datasets from the UCI Machine Learning Repository (Lichman, 2013), presenting different correlation patterns, different degrees of unbalancedness and some of them combining both continuous and categorical variables, will be analyzed under the sparse NB described in Section 3. In the experiments, the performance rates of the classifier shall be estimated according to an $N$ runs $N$-fold cross validation procedure, with $N = 10$. At each fold, the dataset is split into

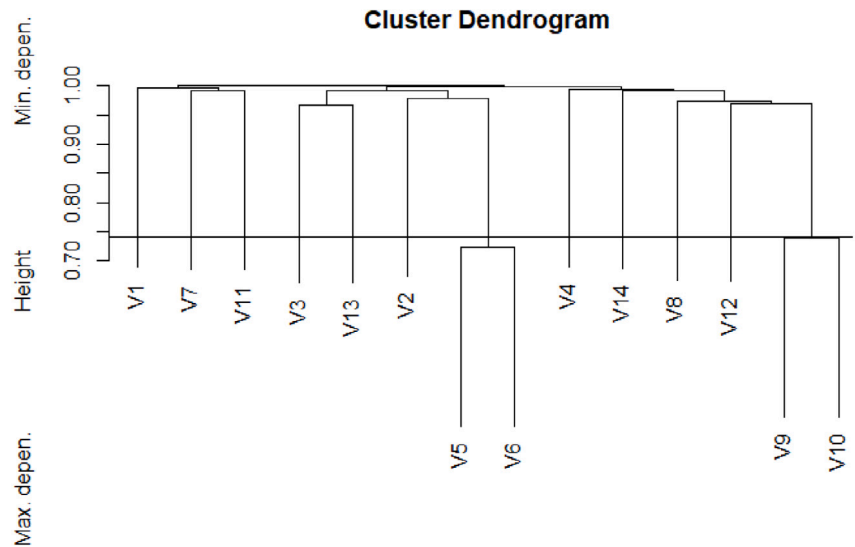

**Fig. 2.** Cluster dendrogram (based on MI correlation) corresponding to the Australian dataset.

---

**Algorithm 1:** Pseudo-code of the sparse NB

1. Select the dependence and the classifier's performance measures.
2. Perform cluster analysis and build the dendrogram.
3. Variable reduction strategy: set specific values for the parameters $\{C, S, q\}$ and initialize $\mathcal{F} = \emptyset$.

**for** $c = 1, \ldots, C$ **do**
    **for** $s = 1, \ldots, \min\{nc(c), S\}$ **do**
        (a) Obtain the $s$th combination of features. For each cluster only one variable is randomly selected with probability $q$, and none with probability $1 - q$.
        (b) Construct the classifier for the $s$th combination of features.
        (c) Evaluate the selected classifier's performance measure and if feasible, add it to $\mathcal{F}$.
    **end**
**end**

4. Variable Selection: select the combination of variables leading to the best performance, among those in $\mathcal{F}$.

---

three sets, the so-called training, validation and testing sets. A tenth of the dataset is used as testing set, and the remaining nine tenths are for training set $\left(\frac{2}{3} \times \frac{9}{10}\right)$ and validation set $\left(\frac{1}{3} \times \frac{9}{10}\right)$. Steps 2, 3(a) and 3(b) of Algorithm 1 are implemented on the training set. The different classifiers built in this way are compared according to their performance results (step 3(c)) on the validation set. The classifier (combination of features) with the highest performance measure on the validation set is chosen, and its average performance rates on the testing set are reported. Special emphasis will be made on the performance behavior and sparsity of the solutions of the proposed method.

### 4.1. Parameters setting

The probability distribution for the features conditioned to the class $X_i \mid C_k$, for $i = 1, \ldots, p$, $k = 1, \ldots, K$, needs to be selected. It is well-known in the literature that the performance of the NB classifier improves when features are categorized using any discretization method (Liu et al., 2002; Boullé, 2004; Boullé, 2006). Therefore, instead of imposing a specific probability distribution (such as the Gaussian), we adopted the discretization method based on an entropy criterion (see Dougherty et al., 1995) and used the `mdlp` routine (Fayyad and Irani, 1993) from the `discretization` package of R.

Now, we discuss the choice of the parameters $\{C, S, q\}$ and the performance criterion.

*Choice of the parameter C*

The value of $C$, which represents the number of cuts in the vertical axis of the dendrogram, is critical for a proper sampling. As a default value, we propose to select the points of the grid where features are clustered. When the routine `hclust` of R is used to generate the dendrogram (as in this work), one has $C = p - 1$, where $p$ is the number of features. In addition, `hclust` specifies where to make the cuts. However, a large value of $C$ may slow down the execution of the algorithm notably and, on the other hand, it may lead to overfitting. For this reason, $C$ will be defined as $\min\{p - 1, 100\}$. As will be seen next, in Section 4.2, such a choice yields a right balance between the performance and the computational time for the considered datasets.

*Choice of the parameter S*

Regarding the value of $S$, which represents the maximum total of combinations for each cut, we tested several possible values for this parameter (see Supplementary Material), and settled on the final choice $S = 25$. Note that under the previous choices of $C$ and $S$ a total of $\max\{25 \times (p - 1), 25 \times 100\}$ combinations of features will be evaluated under the proposed sparse NB in contrast to the total number of possible combinations, equal to $2^p - 1$.

*Choice of the parameter q*

Small values of $q$ are associated with more sparsity (since fewer variables would be included in the combinations to be examined). Therefore, $q$ should be selected in such a way that it provides a compromise between the classifier's performance and the sparsity of the solution. In particular, in the Supplementary Material, different experiments to evaluate how the choice of this parameter affects the results can be found. Here, the selection of this parameter has been addressed according to the dependence matrix $M$, which is defined in Section 3. In particular, when 20% of the matrix elements are higher than 0.1 (that is, from moderate to strong dependence cases), we fix $q = 0.4$ (which implies a sparser solution). Otherwise (few dependent features), $q = 0.6$ will be set.

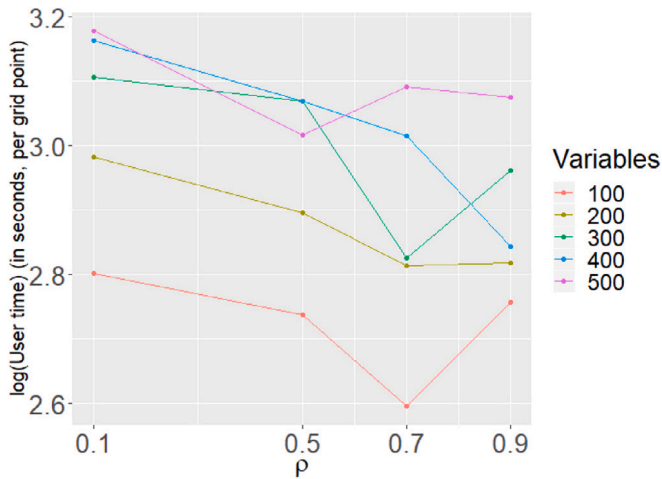

**Fig. 3.** Scalability.

### 4.2. Simulation study

In this section we analyze how the performance of the sparse NB varies as dependence among features increases. In particular, we simulate data following Witten et al. (2014) and according to the model $\mathbf{y} = \mathbf{X}\boldsymbol{\beta} + \boldsymbol{\varepsilon}$ with $p \in \{100, 200, 300, 400, 500\}$. The errors $\varepsilon_1, \ldots, \varepsilon_N$ are iid from a $N(0, 2.5^2)$ distribution. The observations (rows of $\mathbf{X}$) are iid from a $N_p(0, \boldsymbol{\Sigma})$ distribution, where $\boldsymbol{\Sigma}$ is a $p \times p$ block diagonal matrix, with elements as follows:

$$\Sigma_{ij} = \begin{cases} 1 & \text{if } i = j, \\ \rho & \text{if } i \leq \frac{p}{4}, j \leq \frac{p}{4}, i \neq j, \\ \rho & \text{if } \frac{p}{4} + 1 \leq i \leq p, \frac{p}{4} + 1 \leq j \leq p, i \neq j, \\ 0 & \text{otherwise} \end{cases}$$

We explored various values of $\rho$, ranging from 0.1 to 0.9. Furthermore, $\beta_i \sim \text{Unif}[0.9, 1.1]$ for $1 \leq i \leq \lfloor \frac{p}{4} \rfloor$ and $\beta_i \sim \text{Unif}[-\frac{1}{3} - 0.1, -\frac{1}{3} + 0.1]$ otherwise. In other words, there are two sets of $\frac{p}{4}$ and $\frac{3p}{4}$ correlated features, respectively, and all the features are associated with the response. Finally, two classes are defined according to the sign of $y_n$, $n = 1, \ldots, 2000$.

The results in Table 2 have been obtained using the Mutual Information dependence measure (MI), $S = 25$ and values of $q$ fixed as in Section 4.3. Moreover, the performance measure considered for these simulated experiments is the accuracy, and its average performance rates as in (3) on the testing set are reported in Table 2.

Some conclusions can be drawn. On the one hand, in terms of sparsity levels, the sparse NB returns better results in the presence of moderate to strong dependence cases. For datasets where the dependences among features are weak, $\rho = 0.1$, our sparse strategy is able to remove around one third of the total number of variables whereas, in some cases, the ACC is slightly reduced with regards to the classic NB. While $\rho$ increases, our proposal is able to significantly reduce the number of variables considered, achieving better ACC results than the classic NB, as the curse of dependency is alleviated. On the other hand, Fig. 3 reports the logarithm of the average user times (in seconds) when the sparse NB is run on Intel(R) Core(TM) i7-7500U CPU at 2.70 GHz 2.90 GHz with 8.0 GB of RAM. The $X$-axis shows the $\rho$ values whereas each line represents the number of variables of the dataset ($p$). Overall, for weak dependences among features, the behavior of running time is monotonous respect to the number of variables, but this changes when $\rho$ increases.

### 4.3. Datasets and benchmark approaches

The so-called *Breast Cancer Wisconsin (Diagnostic) Data Set, Wine Data Set, Mushroom, Waveform Database Generator Data Set* (version 2), *ISOLET Data Set, Multiple Features Data Set, SPECTF Heart Data Set, German Credit, Page Blocks Classification Data Set* and *Statlog (Australian Credit Approval)* shall be considered. They are described in Table 3, whose first three columns report the dataset name, the number of instances and the class split. The number of continuous variables ($L$) and categorical variables ($L'$) are presented in the last two columns. Three of the ten datasets, *SPECTF Heart Data Set, German Credit* and *Page Blocks Classification Data Set*, are unbalanced datasets, due to the very different sizes of the classes.

We aim to compare the novel method with alternative, well-known strategies for feature selection. In this study, we focus on techniques which perform *hard variable selection* and, in consequence, feature weighting approaches as in Jiang et al. (2019) have not been considered here. There exist two main groups of methods that select features: filters (Guyon et al., 2006; Saeys et al., 2007) and wrappers (Kohavi and John, 1997; Saeys et al., 2007). We selected one filter and one wrapper that are well referenced in the literature and that can be easily adapted to the NB classifier to make a fair comparison. Our choice was the filter CFS introduced in Section 1, see Hall (2000), and the wrapper *Boruta* (Kursa and Rudnicki, 2010). The wrapper *Boruta*, which is in principle designed using a Random Forest strategy, can be modified and adapted to any classifier, in particular, to the NB. These methods are widely spread and can be computed by the routines cfs and Boruta, from R packages FSelector and Boruta, respectively. In order to adapt the wrapper *Boruta* to the NB classifier, we have used the function filterVarImp in R package caret as the function that returns the importance of the attributes, instead of the default getImpRfZ, which is based on the Random Forest classifier. It is important to highlight that the time limit is not an input parameter of the cfs and Boruta routines and therefore, differences in the computational costs were found (to be discussed later). Apart from the previous feature selection methods, that can be applied to any classifier, there are works that specifically deal with variable reduction for the NB. In particular, Boullé (2007) proposes a straightforward Bayesian modern-style approach, the *MAP Approach for Variable Selection* (noted SNB(MAP)), where the conditional probabilities are formulated according to

$$p(C_k|\mathbf{x}) \propto \pi(C_k) \prod_{i=1}^{p} p(x_i|C_k)^{a_i}, \quad k = 1, \ldots, K. \tag{6}$$

In Eq. (6), the values $\{a_i\}_{i=1}^{p}$ are either 1 or 0, depending on whether or not feature $i$ is included in the model. Then, the posterior distribution of the different models (resulting from different choices of $\{a_i\}_{i=1}^{p}$) is evaluated using a shrinkage prior so that parsimonious models are favored. In the same paper, a search heuristic that performs a fast forward backward selection is described and therefore, it has been implemented in this paper to run the different experiments. However, when the number of variables increases, note that the time required to run this method is excessive. For that reason, a time limit of eight hours for the two biggest considered real datasets (*ISOLET* and *Multiple Features*) was fixed. Finally, we have also compared with the Lasso approach for classification (see Vincent and Hansen, 2014), whose goal is precisely the same: obtain good classification performance while selecting few features. The routine fit in R package msgl has been used.

Next, we will break the results down depending on the datasets are balanced or unbalanced. As we will show below, if necessary and motivated by the properties of the dataset, our proposal can be easily adapted in terms of the performance criterion to be optimized and the required constraints on groups of interest. To make a fair comparison, we do not impose any additional constraints and therefore only the performance criterion will change accordingly throughout these sections. However, an illustrative example where constraints are imposed is also included at the end of Section 4.5.

**Table 2**
Average accuracy and sparsity (10 runs 10-fold CV) for simulated datasets.

| p | Method | $\rho = 0.1$ | | $\rho = 0.5$ | | $\rho = 0.7$ | | $\rho = 0.9$ | |
|---|---|---|---|---|---|---|---|---|---|
| | | ACC | *Sparsity* | ACC | *Sparsity* | ACC | *Sparsity* | ACC | *Sparsity* |
| 100 | Sparse NB | 88.20 | 66 | 93.30 | 37 | 93.20 | 24 | 95.05 | 21.30 |
| | Classic NB | 89.50 | 100 | 87.05 | 100 | 87.10 | 100 | 86.95 | 100 |
| 200 | Sparse NB | 90.10 | 113.5 | 93.40 | 52.50 | 95.70 | 30.50 | 96.55 | 20.10 |
| | Classic NB | 90.10 | 200 | 87.30 | 200 | 87.65 | 200 | 86.80 | 200 |
| 300 | Sparse NB | 89.85 | 168.60 | 92.40 | 79.70 | 94.70 | 37.30 | 95.95 | 18.20 |
| | Classic NB | 90.15 | 300 | 87.00 | 300 | 87.15 | 300 | 87.85 | 300 |
| 400 | Sparse NB | 91.90 | 216.30 | 91.45 | 94 | 92.35 | 46.80 | 93.65 | 17 |
| | Classic NB | 89.65 | 400 | 87.25 | 400 | 87.25 | 400 | 87.50 | 400 |
| 500 | Sparse NB | 91.90 | 283.20 | 90.70 | 102.70 | 93.85 | 29.60 | 91.90 | 5.90 |
| | Classic NB | 90.15 | 500 | 87.05 | 500 | 87.45 | 500 | 87.55 | 500 |

**Table 3**
Datasets description.

| Name | Instances | Class split in % | L | L' |
|---|---|---|---|---|
| *Breast Cancer Wisconsin* | 569 | 63(*Benign*)/37(*Malignant*) | 30 | 0 |
| *Wine Data Set* | 178 | 33(*Class 1*)/40(*Class 2*)/27(*Class 3*) | 13 | 0 |
| *Mushroom* | 8124 | 51.8(*edible*)/48.2(*poisonous*) | 0 | 22 |
| *Waveform Database Generator* | 5000 | 33.33(*Class 0*)/33.33(*Class 1*)/33.33(*Class 2*) | 40 | 0 |
| *ISOLET Data Set* | 7797 | 26 equiprobable classes (0.04) | 617 | 0 |
| *Multiple Features* | 2000 | 9 equiprobable classes (0.11) | 649 | 0 |
| *SPECTF Heart* | 267 | 79(*Abnormal*)/21(*Normal*) | 44 | 0 |
| *German Credit* | 1000 | 70(*Class 1*)/30(*Class 2*) | 7 | 13 |
| *Page Blocks Data Set* | 5473 | 90(*Negative*)/10(*Positive*) | 10 | 0 |
| *Australian* | 690 | 55.5(*Negative*)/44.5(*Positive*) | 6 | 8 |

*4.4. Results for balanced datasets*

For comparison purposes, consider the same parameters setting than in Section 4.2, where for *Waveform* dataset, $q$ is equal to 0.6 and, for the remaining balanced databases, $q = 0.4$. We next analyze the performance and sparsity of the method, as well as the running times. The average accuracy, number of variables in the selected combinations and the CPU time in seconds for 1 fold-CV execution are shown by Fig. 4. Moreover, a comparison between the sparse NB with the above-mentioned feature selection methods is made. The results under the classic NB, CFS, *Boruta*, SNB(MAP) and Lasso methods are also shown.

Several conclusions can be drawn at this point. Note that the performance rates under the sparse NB are comparable to the classic NB using between a half and one third of the variables, except for *Waveform Database*. As commented before, the novel approach is intended to address databases with correlated patterns and, for this reason, the outperformance of the sparse NB improves with the dependence among the features. Therefore, since the variables of the *Waveform Database* are almost independent, it is expected that the novel sparse strategy does not yield a significant enhancement in this sense, as Fig. 4 shows.

With regards to the five feature selection methods considered in this study, the next conclusions can be drawn from the figure. Whereas the proposed method achieves competitive ACC and sparsity results, it performs in between the other methods in these two measures. Also, it can be concluded that SNB(MAP) is computationally slower than the sparse NB. In addition, Boruta and CFS are less computationally costly than the sparse NB, but when the number of features increase, it turns out to be exceptionally low.

In summary, it can be deduced that, for balanced datasets with dependencies among the features, the proposed sparse NB leads to a significant reduction in the number of features while keeping the power prediction. Also, it can be concluded that in general, for this kind of datasets, our method and the Lasso seem to achieve the best compromise between accuracy, sparsity and running times.

*4.5. Results for unbalanced datasets*

In this section we deal with three unbalanced datasets. The *SPECTF Heart Data Set*, *German Credit* and *Page Blocks Classification Data Set*, which are unbalanced according to classes. It implies that the use of the ACC, defined by (3), as the performance criterion may not be a sensible choice because of the difference between the classes sizes. Therefore, for these cases, the area under the curve (AUC) as well as the *precision* of the majority class (*Class 1*), calculated by (4), will be the measures to be maximized. The former measure, $precision_1$, leads to good $Recall_2$, since will minimize the *False Class 1*. In addition, the performance at each class, will be inspected via the so-called *Recall*. We have considered the previous two performance measures when selecting the set of variables via sparse NB, and the obtained results are shown in red and blue (respectively) in Fig. 5. Finally, $q = 0.6$ for *German Credit* and *Page Blocks*, whereas is equal to 0.4 in the case of *SPECTF Heart Data Set*.

Again, the performance results, the sparsity results and the running times are reported in Fig. 5. For each dataset, two graphics are shown. The images on the left represent the AUC versus the sparsity, while the *Recall* of the majority and minority classes ($Recall_1$ and $Recall_2$, respectively) are drawn on the right side. Note that the Boruta results for *SPECTF* database are not reported since this dataset does not satisfy the technical requirements of the implementation of that method.

The performance rates under the sparse NB are comparable to the results obtained with all the features, since the AUC (respectively, the *precision*) has been used as performance criterion and the novel approach keeps at least the area under the curve (or *precision*) obtained by the classic NB. The sparse NB is able to reduce to less than half the number of variables in the case of *SPECTF Heart* dataset; it removes one fourth of the variables of *German* dataset and one third in *Page Blocks Data Set*. Now, if we compare to CFS, *Boruta*, SNB(MAP) and the Lasso, it can be observed how, although they tend to be sparser, they increase significantly the misclassification rate on the minority class ($Recall_2$), since, in general, they tend to increase the correct classification for the majority class ($Recall_1$) and to decrease the minority one. The latest results assert the need to choose an appropriate performance measure according to the properties of the dataset.

Therefore, with regards to the unbalanced databases, the sparse NB provides more balanced *Recall* values, in the sense that the performance of the least frequent class is not so reduced. Another illustration is given by Table 4, where the *Australian Credit Approval* is considered. As commented before, in this case, the positive class (the load is granted) is the most risky. For these cases, the sparse NB would be the most suitable choice, not only because the performance criterion to be used can be easily adapted but also because while optimizing such criterion, constraints on acceptable performance measures can be included. The second row of Table 4 shows the results for the Sparse NB if the ACC is considered as performance criterion and no additional constraints are imposed. However, the ACC can be optimized whereas a performance constraint on the *Recall* of the positive class is considered (*Recall +* > 85), as can be seen in the third row of Table 4. As a final example, we are interested in maximizing the *Recall +* instead. Note that the improvement in the positive class will be at the expense of reducing the *Recall* - and therefore admissible values for it have been imposed via a threshold value to avoid worsening it, say *Recall* - > 60 (last row).

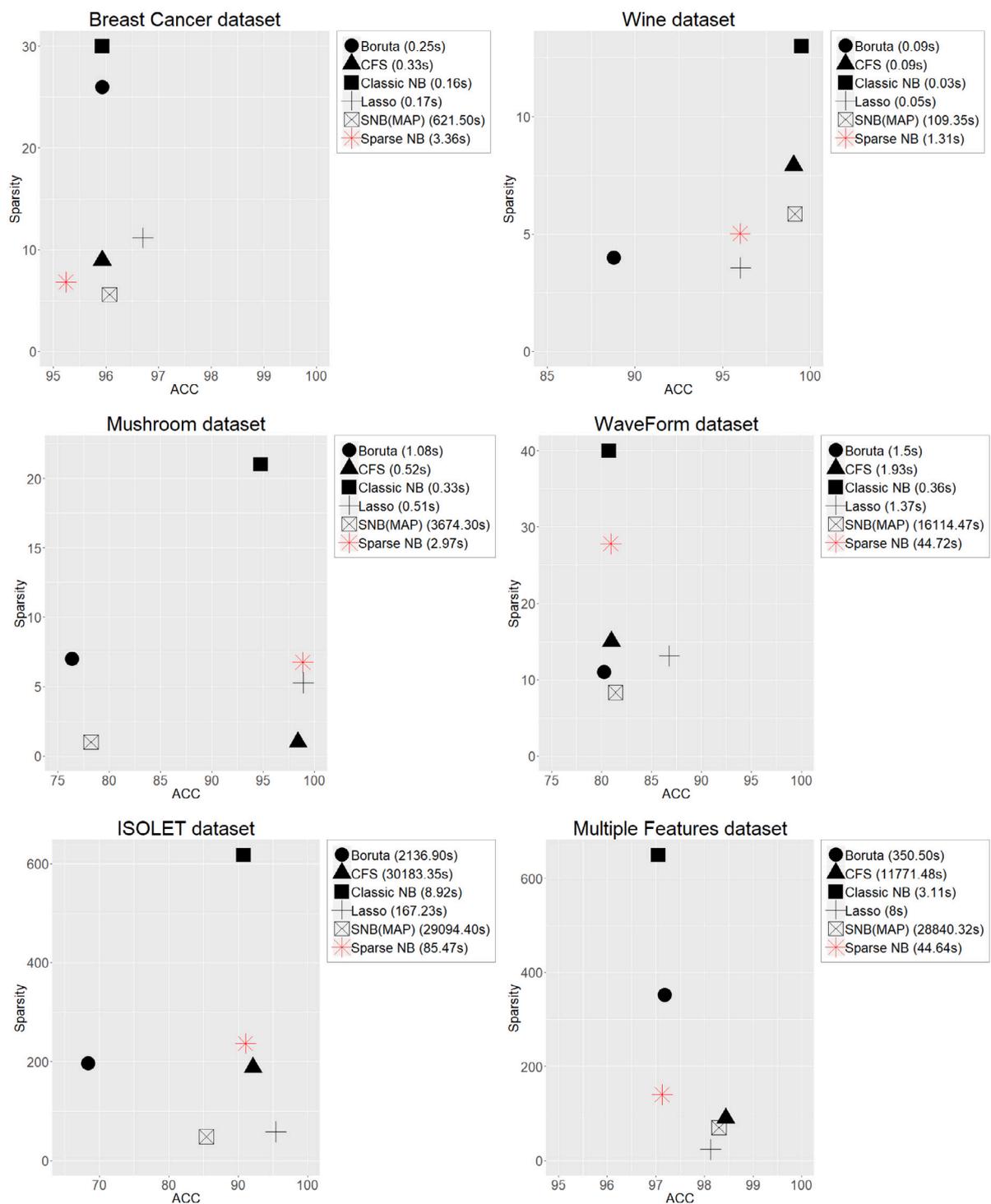

**Fig. 4.** Average accuracy, sparsity and CPU time (10 runs 10-fold CV) for *Breast Cancer* (*BC*), *Wine*, *Mushroom*, *Waveform*, *ISOLET* and *Multiple Features* (*Mult. Feat.*) datasets.

**Table 4**
Average performance and sparsity (10 runs 10-fold CV) for *Australian* dataset using the sparse NB with different performance measures to select the set of variables.

| Method | Recall − | Recall + | ACC | Sparsity |
|---|---|---|---|---|
| Classic NB | 91.10 | 78.78 | 85.61 | 14 |
| Sparse NB (ACC) | 84.59 | 85.83 | 85.15 | 5.78 |
| Sparse NB (ACC); *Recall +* > 85 | 84.14 | 86.48 | 85.19 | 5.53 |
| Sparse NB (*Recall +*); *Recall −* > 60 | 79.93 | 92.35 | 85.46 | 1.4 |

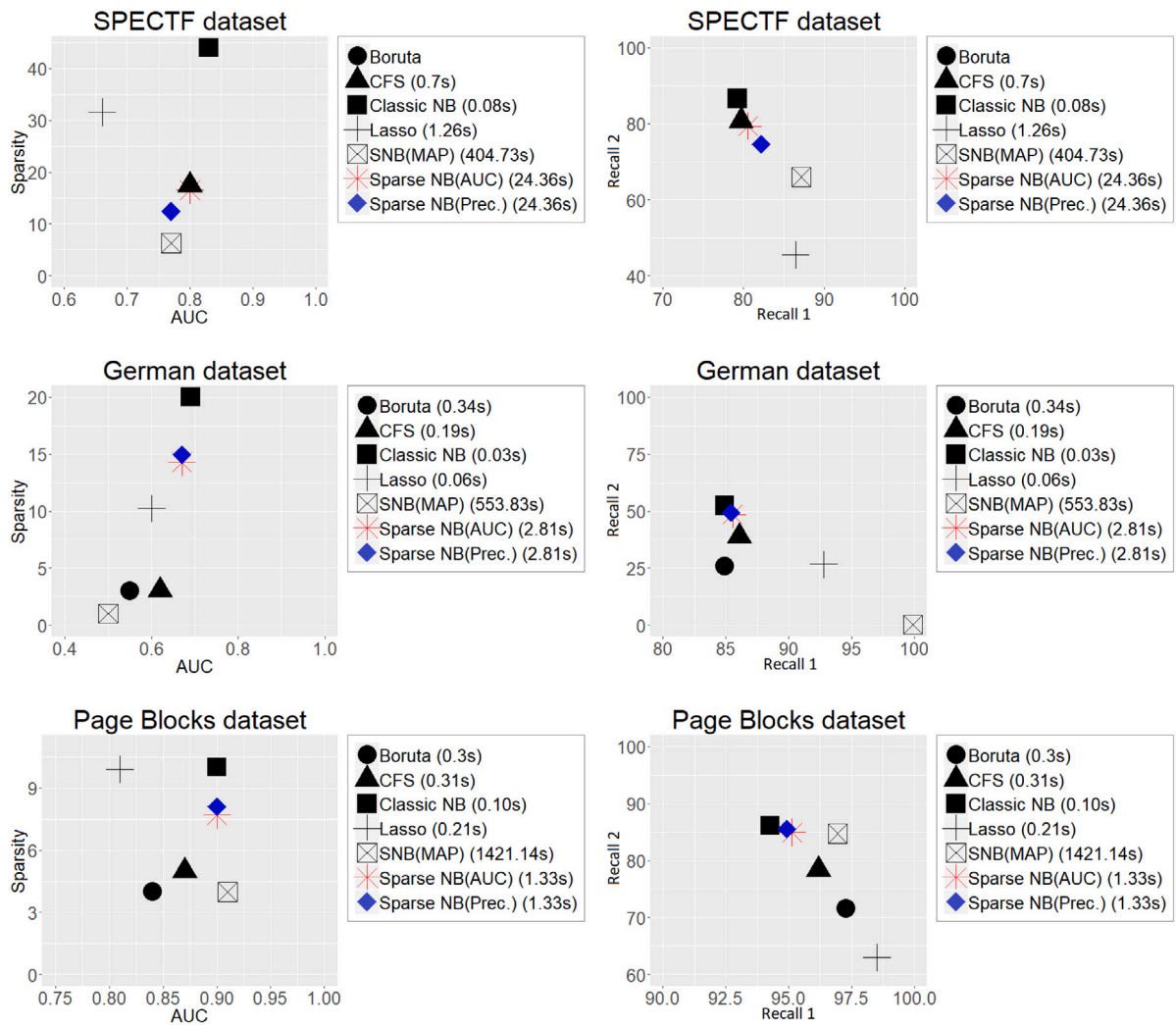

**Fig. 5.** Average performance, sparsity and CPU time (10 runs 10-fold CV) for *SPECTF*, *German* and *Page Blocks* datasets.

To sum up, for unbalanced datasets with dependent variables, the considered benchmark methods tend to be sparser than our approach but at the cost of damaging unpredictably the performance of the classifier and, in particular, the *Recall* of the least frequent class. In contrast, the novel method allows the user to set the performance measure that best suits it as well as admissible values for specific performance measures, which turns out advantageous for unbalanced datasets or for cases in which misclassification costs are strongly class-dependent.

## 5. Conclusions and extensions

In this paper, a new version of the Naïve Bayes (NB) classifier for dealing with datasets with correlated patterns is proposed with the aim of improving the sparsity of the solution. In order to achieve sparsity, a variable reduction technique is embedded into the classifier. Such a variable reduction strategy is based on clustering the features in terms of their dependence degree, and it selects combinations of features that, being as independent as possible, lead to a good performance rate. The performance measure used in the algorithm can be given by the out-of-sample accuracy, or more generally, an estimate of the expected misclassification cost, among others. The proposed methodology has been tested on synthetic datasets and ten real datasets of different sizes and properties. The numerical results show that not only sparse solutions are attained, but also the performance rates are comparable or better than those achieved under the classic version of the NB, where all features are taken into account for classifying. In addition, when compared with benchmark approaches, the novel method turns out especially advisable for datasets where the classes are unbalanced and/or of different importance. This fact stems from the flexibility of our method in the selection of the performance measure and the ability to include constraints on certain performance measures for feature selection, which does not occur with the feature selection approaches proposed in the literature.

In this work, sparsity has been explored in the case of the NB because of its tractability and good performance, but other classifiers could have also been tested instead, as for example the support vector machines. Work on these issues is underway.

**CRediT authorship contribution statement**

**Rafael Blanquero:** Conceptualization, Methodology, Writing - original draft, Writing - review & editing, Supervision. **Emilio Carrizosa:** Conceptualization, Methodology, Writing - original draft, Writing - review & editing, Supervision. **Pepa Ramírez-Cobo:** Conceptualization, Methodology, Writing - original draft, Writing - review & editing, Supervision. **M. Remedios Sillero-Denamiel:** Conceptualization,

Methodology, Software, Validation, Writing - original draft, Writing - review & editing.

**Acknowledgments**

This research is partially supported by research grants and projects MTM2015-65915-R (Ministerio de Economía y Competitividad, Spain) and PID2019-110886RB-I00 (Ministerio de Ciencia, Innovación y Universidades, Spain), FQM-329 and P18-FR-2369 (Junta de Andalucía, Spain), PR2019-029 (Universidad de Cádiz, Spain), Fundación BBVA and EC H2020 MSCA RISE NeEDS Project (Grant agreement ID: 822214). This support is gratefully acknowledged.

**Appendix A. Supplementary material**

Supplementary material related to this article can be found online at https://doi.org/10.1016/j.cor.2021.105456.

**References**

"ann" Ratanamahatana, C., Gunopulos, D., 2003. Feature selection for the naive bayesian classifier using decision trees. Appl. Artif. Intell. 17 (5–6), 475–487.

Benati, S., García, S., 2014. A mixed integer linear model for clustering with variable selection. Comput. Oper. Res. 43, 280–285.

Benítez-Peña, S., Blanquero, R., Carrizosa, E., Ramírez-Cobo, P., 2019. On support vector machines under a multiple-cost scenario. Adv. Data Anal. Classif. 13 (3), 663–682.

Bermejo, P., Gámez, J.A., Puerta, J.M., 2014. Speeding up incremental wrapper feature subset selection with Naive Bayes classifier. Knowl.-Based Syst. 55, 140–147.

Blanquero, R., Carrizosa, E., Jiménez-Cordero, A., Martín-Barragán, B., 2020. Selection of time instants and intervals with support vector regression for multivariate functional data. Comput. Oper. Res. 123, 105050.

Blanquero, R., Carrizosa, E., Molero-Río, C., Romero Morales, D., 2021a. Optimal randomized classification trees. Comput. Oper. Res. 132, 105281.

Blanquero, R., Carrizosa, E., Ramírez-Cobo, P., Sillero-Denamiel, M.R., 2021b. A cost-sensitive constrained lasso. Adv. Data Anal. Classif. 15, 121–158.

Boullé, M., 2004. Khiops: A statistical discretization method of continuous attributes. Mach. Learn. 55 (1), 53–69.

Boullé, M., 2006. MODL: A Bayes optimal discretization method for continuous attributes. Mach. Learn. 65 (1), 131–165.

Boullé, M., 2007. Compression-based averaging of selective naive Bayes classifiers. J. Mach. Learn. Res. 8, 1659–1685.

Cai, A., Tsay, R., Chen, R., 2009. Variable selection in linear regression with many predictors. J. Comput. Graph. Statist. 18 (3), 573–591.

Carrizosa, E., Guerrero, V., 2014. Biobjective sparse principal component analysis. J. Multivariate Anal. 132, 151–159.

Carrizosa, E., Molero-Río, C., Romero Morales, D., 2021. Mathematical optimization in classification and regression trees. TOP 29, 5–33. http://dx.doi.org/10.1007/s11750-021-00594-1.

Carrizosa, E., Nogales-Gómez, A., Romero Morales, D., 2016. Strongly agree or strongly disagree?: Rating features in support vector machines. Inform. Sci. 329, 256–273.

Carrizosa, E., Olivares-Nadal, A.V., Ramírez-Cobo, P., 2017. A sparsity-controlled vector autoregressive model. Biostatistics 18 (2), 244–259.

Carrizosa, E., Romero Morales, D., 2013. Supervised classification and mathematical optimization. Comput. Oper. Res. 40 (1), 150–165.

Chen, S., Webb, G.I., Liu, L., Ma, X., 2020. A novel selective naïve Bayes algorithm. Knowl.-Based Syst. 192, 105361.

Domingos, P., Pazzani, M., 1996. Beyond independence: Conditions for the optimality of the simple Bayesian classifier. In: Proceedings of the Thirteenth International Conference on Machine Learning. Morgan Kaufmann, pp. 105–112.

Domingos, P., Pazzani, M., 1997. On the optimality of the simple Bayesian classifier under zero-one loss. Mach. Learn. 29 (2–3), 103–130.

Dougherty, J., Kohavi, R., Sahami, M., 1995. Supervised and unsupervised discretization of continuous features. In: Prieditis, A., Russell, S. (Eds.), Machine Learning Proceedings 1995. pp. 194–202.

Fayyad, U.M., Irani, K.B., 1993. Multi-interval discretization of continuous valued attributes for classification learning. In: Proceedings of the 13th International Joint Conference on Artificial Intelligence. Morgan-Kaufmann, pp. 1022–1029.

Feng, G., Guo, J., Jing, B.-Y., Sun, T., 2015. Feature subset selection using naive Bayes for text classification. Pattern Recognit. Lett. 65, 109–115.

George, E., McCulloch, R., 1993. Variable selection via gibbs sampling. J. Amer. Statist. Assoc. 88 (423), 881–889.

Guan, G., Guo, J., Wang, H., 2014. Varying Naïve Bayes models with applications to classification of chinese text documents. J. Bus. Econom. Statist. 32 (3), 445–456.

Guyon, I., Gunn, S., Nikravesh, M., Zadeh, L.A., 2006. Feature Extraction. Foundations and Applications. In: Studies in Fuzziness and Soft Computing, vol. 207, Springer.

Hall, M.A., 2000. Correlation-based feature selection for discrete and numeric class machine learning. In: Proceedings of the Seventeenth International Conference on Machine Learning. ICML '00, Morgan Kaufmann Publishers Inc., San Francisco, CA, USA, pp. 359–366.

Hand, D.J., Yu, K., 2001. Idiot's Bayes - not so stupid after all? Internat. Statist. Rev. 69 (3), 385–398.

Hastie, T., Tibshirani, R., Friedman, J., 2001. The Elements of Statistical Learning. Springer, NY.

Hastie, T., Tibshirani, R., Wainwright, M., 2015. Statistical Learning with Sparsity. The Lasso and Generalizations. CRC Press.

Hoeffding, W., 1948. A non-parametric test of independence. Ann. Math. Stat. 19 (4), 546–557.

Jiang, L., Cai, Z., Zhang, H., Wang, D., 2012. Not so greedy: Randomly selected naive Bayes. Expert Syst. Appl. 39 (12), 11022–11028.

Jiang, L., Zhang, H., Cai, Z., Su, J., 2005. Evolutional naive Bayes. In: Proceedings of the 1st International Symposium on Intelligent Computation and Its Applications. ISICA 2005, China University of Geosciences Press, pp. 344–350.

Jiang, L., Zhang, L., Li, C., Wu, J., 2019. A correlation-based feature weighting filter for naive Bayes. IEEE Trans. Knowl. Data Eng. 31 (2), 201–213.

Jiang, L., Zhang, L., Yu, L., Wang, D., 2019. Class-specific attribute weighted naive Bayes. Pattern Recognit. 88, 321–330.

Kinney, J.B., Murugan, A., Callan, C.G., Cox, E.C., 2010. Using deep sequencing to characterize the biophysical mechanism of a transcriptional regulatory sequence. Proc. Natl. Acad. Sci. 107 (20), 9158–9163.

Kohavi, R., John, G.H., 1997. Wrappers for feature subset selection. Artificial Intelligence 97 (1–2), 273–324.

Kuncheva, L.I., 2006. On the optimality of Naïve Bayes with dependent binary features. Pattern Recognit. Lett. 27 (7), 830–837.

Kursa, M.B., Rudnicki, W.R., 2010. Feature selection with the boruta package. J. Stat. Softw. 36 (11), 1–13.

Langley, P., Sage, S., 1994. Induction of selective Bayesian classifiers. In: Proceedings of the Tenth International Conference on Uncertainty in Artificial Intelligence. pp. 399–406.

Leevy, J.L., Khoshgoftaar, T.M., Bauder, R.A., Seliya, N., 2018. A survey on addressing high-class imbalance in big data. J. Big Data 5 (42), http://dx.doi.org/10.1186/S40537-018-0151-6.

Lichman, M., 2013. UCI Machine Learning Repository. University of California, School of Information and Computer Sciences, Irvine.

Lin, D., Foster, D.P., Ungar, L.H., 2011. VIF regression: A fast regression algorithm for large data. J. Amer. Statist. Assoc. 106 (493), 232–247.

Linfoot, E., 1957. An informational measure of correlation. Inf. Control 1 (1), 85–89.

Liu, H., Hussain, F., Tan, C.L., Dash, M., 2002. Discretization: An enabling technique. Data Min. Knowl. Discov. 6 (4), 393–423.

Maldonado, S., Carrizosa, E., Weber, R., 2015. Kernel Penalized K-means: A feature selection method based on Kernel K-means. Inform. Sci. 322, 150–160.

McCallum, A., Nigam, K., 1998. A comparison of event models for naive bayes text classification. In: AAAI-98 Workshop on Learning for Text Categorization, vol. 752. pp. 41–48.

Minnier, J., Yuan, M., Liu, J.S., Cai, T., 2015. Risk classification with an adaptive naive Bayes kernel machine model. J. Amer. Statist. Assoc. 110 (509), 393–404.

Mukherjee, S., Sharma, N., 2012. Intrusion detection using naive Bayes classifier with feature reduction. Proc. Technol. 4, 119–128.

R Core Team, 2017. R: A Language and Environment for Statistical Computing. R Foundation for Statistical Computing, Vienna, Austria.

Reshef, D.N., Reshef, Y.A., Finucane, H.K., Grossman, S.R., McVean, G., Turnbaugh, P.J., Lander, E.S., Mitzenmacher, M., Sabeti, P.C., 2011. Detecting novel associations in large data sets. Science 334 (6062), 1518–1524.

Rezaei, M., Cribben, I., Samorani, M., 2018. A clustering-based feature selection method for automatically generated relational attributes. Ann. Oper. Res. http://dx.doi.org/10.1007/s10479-018-2830-2.

Saeys, Y., Inza, I., Larrañaga, P., 2007. A review of feature selection techniques in bioinformatics. Bioinformatics 23 (19), 2507–2517.

Sharpee, T., Rust, N.C., Bialek, W., 2004. Analyzing neural responses to natural signals: Maximally informative dimensions. Neural Comput. 16 (2), 223–250.

Sokolova, M., Lapalme, G., 2009. A systematic analysis of performance measures for classification tasks. Inf. Process. Manage. 45 (4), 427–437.

Székely, G.J., Rizzo, M.L., Bakirov, N.K., 2007. Measuring and testing dependence by correlation of distances. Ann. Statist. 35 (6), 2769–2794.

Tang, B., He, H., Baggenstoss, P.M., Kay, S., 2016a. A Bayesian classification approach using class-specific features for text categorization. IEEE Trans. Knowl. Data Eng. 28 (6), 1602–1606.

Tang, B., Kay, S., He, H., 2016b. Toward optimal feature selection in naive Bayes for text categorization. IEEE Trans. Knowl. Data Eng. 28 (9), 2508–2521.

Turhan, B., Bener, A., 2009. Analysis of Naive Bayes' assumptions on software fault data: An empirical study. Data Knowl. Eng. 68 (2), 278–290.

Vincent, M., Hansen, N.R., 2014. Sparse group lasso and high dimensional multinomial classification. Comput. Statist. Data Anal. 71, 771–786.

Witten, D.M., Shojaie, A., Zhang, F., 2014. The cluster elastic net for high-dimensional regression with unknown variable grouping. Technometrics 56 (1), 112–122.

Wolfson, J., Bandyopadhyay, S., Elidrisi, M., Vazquez-Benitez, G., Vock, D.M., Musgrove, D., Adomavicius, G., Johnson, P.E., O'Connor, P.J., 2015. A Naive Bayes machine learning approach to risk prediction using censored, time-to-event data. Stat. Med. 34 (21), 2941–2957.

Zhang, H., 2004. The optimality of Naive Bayes. In: Barr, V., Markov, Z. (Eds.), Proceedings of the Seventeenth International Florida Articial Intelligence Research Society Conference. pp. 562–567.

Zhang, H., Jiang, L., Yu, L., 2020. Class-specific attribute value weighting for Naive Bayes. Inform. Sci. 508, 260–274.

Zhang, M., Peña, J., Robles, V., 2009. Feature selection for multi-label naive Bayes classification. Inform. Sci. 179 (456), 3218–3229.

Zou, H., Hastie, T., 2005. Regularization and variable selection via the elastic net. J. R. Stat. Soc. Ser. B Stat. Methodol. 67 (2), 301–320.